\documentclass{article}

\usepackage{arxiv}

\usepackage[utf8]{inputenc} 
\usepackage[T1]{fontenc}    
\usepackage{url}            
\usepackage{booktabs}       
\usepackage{amsfonts}       
\usepackage{nicefrac}       
\usepackage{microtype}      
\usepackage{lipsum}
\usepackage{graphicx}

\usepackage{amsmath}
\usepackage{listings}
\usepackage{xcolor}
\usepackage{multirow}
\usepackage{colortbl}
\usepackage{arydshln}
\usepackage{bm}

\makeatletter
\def\adl@drawiv#1#2#3{%
        \hskip.5\tabcolsep
        \xleaders#3{#2.5\@tempdimb #1{1}#2.5\@tempdimb}%
                #2\z@ plus1fil minus1fil\relax
        \hskip.5\tabcolsep}
\newcommand{\cdashlinelr}[1]{%
  \noalign{\vskip\aboverulesep
           \global\let\@dashdrawstore\adl@draw
           \global\let\adl@draw\adl@drawiv}
  \cdashline{#1}
  \noalign{\global\let\adl@draw\@dashdrawstore
           \vskip\belowrulesep}}
\makeatother

\definecolor{align}{RGB}{0, 102, 0}
\definecolor{diversify}{RGB}{153, 76, 0}
\definecolor{minimize}{RGB}{0, 0, 153}

\definecolor{deepblue}{rgb}{0,0,0.8}
\definecolor{deepgreen}{rgb}{0,0.5,0}

\lstdefinestyle{mystyle}{
    language=Python, 
    basicstyle=\ttfamily\scriptsize, 
    numbers=left, 
    numberstyle=\tiny,
    frame=single, 
    breaklines=true, 
    commentstyle=\color{deepblue}, 
    keywordstyle=\color{black}, 
    identifierstyle=\color{black}, 
}

\title{Align, Minimize and Diversify: A Source-Free Unsupervised Domain Adaptation Method for Handwritten Text Recognition}

\author{
María Alfaro-Contreras and Jorge Calvo-Zaragoza \\
Pattern Recognition and Artificial Intelligence Group \\
University of Alicante\\
\texttt{\{malfaro, jcalvo\}@dlsi.ua.es} \\
}

\begin{document}
\maketitle
\begin{abstract}
This paper serves to introduce the Align, Minimize and Diversify (AMD) method, a Source-Free Unsupervised Domain Adaptation approach for Handwritten Text Recognition (HTR). This framework decouples the adaptation process from the source data, thus not only sidestepping the resource-intensive retraining process but also making it possible to leverage the wealth of pre-trained knowledge encoded in modern Deep Learning architectures. Our method explicitly eliminates the need to revisit the source data during adaptation by incorporating three distinct regularization terms: the \emph{Align} term, which reduces the feature distribution discrepancy between source and target data, ensuring the transferability of the pre-trained representation; the \emph{Minimize} term, which encourages the model to make assertive predictions, pushing the outputs towards one-hot-like distributions in order to minimize prediction uncertainty, and finally, the \emph{Diversify} term, which safeguards against the degeneracy in predictions by promoting varied and distinctive sequences throughout the target data, preventing informational collapse. Experimental results from several benchmarks demonstrated the effectiveness and robustness of AMD, showing it to be competitive and often outperforming DA methods in HTR.
\end{abstract}


\begin{figure}[!ht]
    \centering
    \includegraphics[width=1\linewidth]{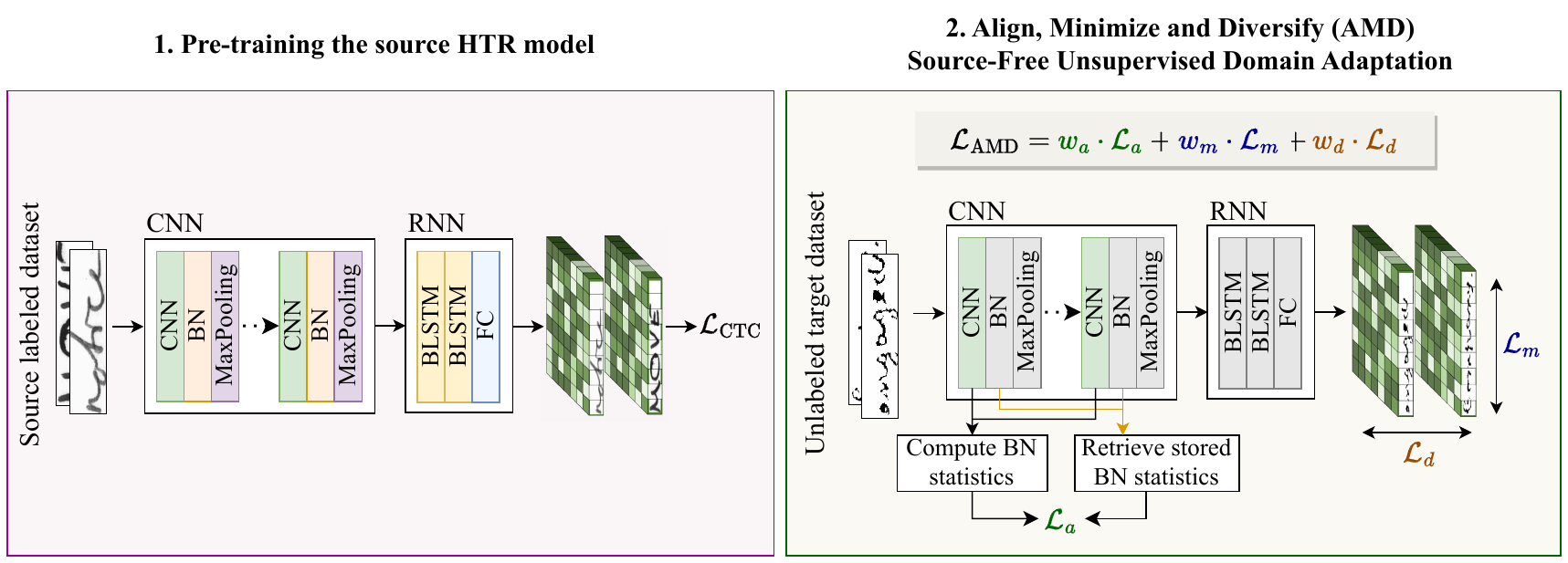}
    \caption{Scheme of the proposed Align, Minimize and Diversify (AMD) method. The approach consists of two stages: (i) pre-training the HTR model using a labeled source dataset, and (ii) adapting it to recognize a new target domain using only its images in an unsupervised manner. AMD uses three loss terms to fine-tune the pre-trained source model: (i) the \textbf{\textcolor{align}{Align}} term, $\mathcal{L}_a$, to align source and target graphical feature distributions; (ii) the \textbf{\textcolor{minimize}{Minimize}} term, $\mathcal{L}_m$, to guide the frame-wise predictions towards one-hot-like vectors, and (iii) the \textbf{\textcolor{diversify}{Diversify}} term, $\mathcal{L}_d$, to ensure diverse sequences throughout the target data. $w_a$, $w_m$, and $w_d$ are hyper-parameters that control the importance of each term in the loss. The layers shaded in gray in the second stage remain fixed during the adaptation.}
    \label{fig:proposal_scheme}
\end{figure}

\section{Introduction}
\label{sec:introduction}

Handwritten Text Recognition (HTR) is an important research field within computer vision. HTR systems are designed to convert handwritten text images into editable and searchable digital text, thus facilitating the extraction of valuable information from vast repositories of handwritten documents~\cite{Muehlberger:JD:2019}. HTR models are typically trained on specific datasets representing particular domains. While these approaches have proven to perform successfully with well-defined datasets in which training and test data are drawn from the same distribution or domain, deploying HTR in real-world scenarios is often a significant challenge owing to the inherent diversity and variability in handwriting styles from different sources.

One critical challenge in HTR is the recognition of out-of-domain (OOD) samples, which refers to text instances that deviate significantly from the data distribution encountered when training the model (i.e., the training or source domain). This problem hinders the generalization capability of conventional HTR models, thus rendering them less effective when confronted with previously unseen handwriting styles~\cite{Zhang:CVPR:2011, Zhang:TPMAI:2012, Soullard:ICDAR:2019, Bhunia:CVPR:2021}. It is necessary to address this issue in order to extend the applicability of HTR systems to a broader spectrum of real-world scenarios in which OOD samples are prevalent.

The objective of this work is to mitigate the OOD challenge by introducing a new approach for Source-Free Unsupervised Domain Adaptation (SFUDA) in HTR. This approach, denominated as Align, Minimize and Diversify (AMD), allows a trained HTR model to adapt to new domains using only the target domain images, without the need for their labels or access to the data used when training the source model (see Fig.~\ref{fig:proposal_scheme}). This source-free approach is particularly relevant given the current wealth of pre-trained models, and enables their adaptation in cases in which the resources required in order to re-train them might not exist (owing to constraints such as time, computational resources or data availability), unlike that which occurs in classical Domain Adaptation (DA).

The contributions of this work are threefold: (i) the proposal of a new three-term training objective for source-free AMD model adaptability; (ii) comprehensive experimentation encompassing 16 distinct source-target configurations that span both real-world and synthetic datasets, and (iii) the demonstration of noticeable improvements to model performance in all the scenarios considered.

\section{Related work}
\label{sec:background}

\paragraph{\textbf{Handwritten Text Recognition.}} Text recognition has evolved from traditional Optical Character Recognition (OCR) methods to the current era of Deep Learning (DL). Early techniques relied on feature engineering and conventional OCR algorithms~\cite{Mori:OCR:1999}, gradually paving the way toward the paradigm-shifting influence of DL architectures, such as Convolutional Recurrent Neural Networks (CRNN)~\cite{Shi:TPAMI:2017} and the Transformer~\cite{Li:AAAI:2023}. Although the latter has recently received attention in HTR~\cite{Parres:ICDAR:2023}, the former, when combined with the Connectionist Temporal Classification (CTC) loss function~\cite{Graves:ICML:2006}, is still the most prominent architecture~\cite{Puigcerver:ICDAR:2017, puigcerver2018pylaia, Cascianelli:IJDAR:2022, Davoudi:NCA:2023, Kang:PR:2022, Vidal:PR:2023} owing to its superior performance over the Transformer~\cite{Kang:PR:2022}.

\paragraph{\textbf{Domain Adaptation.}} As HTR systems transitioned from controlled environments to real-world applications, the challenge of accommodating OOD samples emerged. The inability of conventional models to robustly generalize the cases of varying handwriting styles and historical contexts prompted an interest in DA methodologies tailored to the intricacies of HTR~\cite{Zhang:CVPR:2019, Kang:WACV:2020, Zhang:TIP:2021, Tang:ICASSP:2022}. These approaches, while being relatively effective, are constrained by the need to access annotated data from the source domain during adaptation. In this work, we bypass this data constraint and instead leverage the unlabeled target images directly, employing a three-term regularization loss for adaptation. Our aim is to further extend the capabilities of HTR systems in scenarios in which the original annotated source data is unavailable or inaccessible.

\section{Methodology}
\label{sec:methodology}

This work tackles the problem of SFUDA for HTR. This framework addresses the challenge of adapting an HTR model to a target domain without having access to the original source data on which the model was initially trained. The model, which has learned feature representations in the source domain, must adjust to the new, unlabeled data in the target domain. The objective is to overcome the OOD challenge without the need to revisit the source data. The following sections break down the two different stages of the problem: (i) pre-training the source HTR model, and (ii) the proposed AMD method with which to adapt such a model in order to recognize the target domain.

\subsection{Handwritten Text Recognition model}
\label{subsec:htr_model}
HTR has the objective of retrieving the series of characters that appear in a text image. The recognition problem is, therefore, a sequence labeling task~\cite{Graves:SL:2012}.

Formally, let $\mathcal{X}$ represent a space of text images and $\Sigma^{*}$ represent the complete set of possible corresponding sequences where $\Sigma$ represents the character alphabet. In a supervised sequence labeling scheme, we assume the existence of a set $\mathcal{T} = \left\{\left(x_{i},\mathbf{z}_{i}\right) : x_{i}\in\mathcal{X},\;\mathbf{z}_{i}\in\Sigma^{*}\right\}_{i=1}^{|\mathcal{T}|}$ that associates a given image $x_{i}$ with the sequence of characters $\mathbf{z}_{i} = \left(z_{i1},z_{i2},\ldots,z_{iN}\right)$ contained within it. We hypothesize that an underlying function $f : \mathcal{X} \rightarrow \Sigma^{*}$ relates a given text image $x_{i}\in\mathcal{X}$ to its corresponding sequence of characters $\mathbf{z}_{i} \in \Sigma^{*}$.  

In order to approximate the $f$ function as $\hat{f}$, CRNN trained with the CTC loss function represents a state-of-the-art approach~\cite{puigcerver2018pylaia, Cascianelli:IJDAR:2022, Davoudi:NCA:2023, Vidal:PR:2023}. The convolutional block extracts relevant features from the input text image, while the recurrent layers interpret these features as sequences of characters. The CTC loss function enables training without requiring explicit information about the location of the characters in the image thanks to the inclusion of an additional ``\emph{blank}'' symbol in the alphabet, i.e., $\Sigma' = \Sigma \cup \left\{\textrm{\textit{blank}}\right\}$~\cite{Graves:ICML:2006}.

During the prediction phase, the CTC loss function assumes that the CRNN architecture contains a fully-connected network with $|\Sigma'|$ outputs and a softmax activation function. In this context, assuming that for a given input image $x$, the recurrent block outputs sequences of length $K$, this process retrieves the most probable character per frame as described in Eq.~\ref{eq:CTC_greedy_decoding}:

\begin{equation}
\label{eq:CTC_greedy_decoding}
    \hat{\bm{\pi}} = \arg\max_{\bm{\pi}\in\Sigma'^{K}}{\prod_{k=1}^{K}{y^{\pi_{k}}_{x_k}}}
\end{equation}
\noindent where $y^{\pi_{k}}_{x_k}$ represents the activation probability of character $\pi_{k}$ at frame $k$, and $\hat{\bm{\pi}}$ denotes the retrieved sequence of length $K$. 

$\hat{\bm{\pi}}$ can be converted into actual text in several ways, with a \emph{greedy} decoding policy being that typically considered. This consists of a mapping function $\mathcal{B}\left(\cdot\right)$, which merges consecutive repeated characters and removes the \textit{blank} symbol. The predicted sequence is, therefore, obtained as $\hat{\mathbf{z}} = \mathcal{B}\left(\hat{\bm{\pi}}\right)$, where $\left|\hat{\mathbf{z}}\right| \leq K$.

\subsection{Problem statement}
\label{subsec:problem_statement}

In an SFUDA scenario, we have access only to the unlabeled target text images, $\mathcal{X}_{T}$, and the pre-trained HTR source model, $\mathcal{M}_{S}$, because the source domain dataset, $\mathcal{T}_{S}$, is no longer available. Moreover, the target text images originate from a different marginal distribution: the independent and identically distributed (i.i.d.) assumption is violated.\footnote{Note that $\Sigma_S$ and $\Sigma_T$ do not necessarily have to be the same, but they are not necessarily different either. This is analogous to Partial DA~\cite{Zhang:CVPR:2018, Cao:ECCV:2018, Kundu:CVPR:2020}. In our experiments, we have a minimum overlap degree of 74\%. The precise degree of overlap for each source-target configuration can be found in the ``Supplementary Material''} Our objective is, therefore, to develop a model that performs well on the target text images by satisfying the following properties:
\begin{enumerate}
    \item \textbf{It must align the target data distribution with that of the source data.} The learned feature space is conditioned by the source data and it is in this space that the model has learned to solve the HTR task. It is necessary to reduce the discrepancy between source and target features---i.e., to overcome the domain shift---in order to ensure a proper knowledge transfer.

    \item \textbf{It must output confident predictions.} Solely aligning the feature distributions does not guarantee a correct performance in the target domain. It is, therefore, necessary to minimize uncertainty at the output such that target predictions are similar to one-hot vectors at the frame level.

    \item \textbf{It must prevent an informational collapse.} Considering only the two properties mentioned above, a trivial solution would be to assign the same sequence to all unlabeled target data. However, degenerate predictions of this nature must be avoided by ensuring that the target outputs are both individually certain and globally diverse.
\end{enumerate}

To this end, we introduce AMD (Align, Minimize and Diversify), a SFUDA method for the fine-tuning of pre-trained HTR models using only unlabeled target data. Fig.~\ref{fig:proposal_scheme} shows an overview of our approach.

\subsection{Align, Minimize and Diversify (AMD)}
\label{subsec:amd_method}

We propose a new loss function with three terms, each of which is dedicated to dealing with one of the aforementioned properties. These are described in detail below.

\subsubsection{Align.}
\label{subsubsec:align}

Since we have access only to the pre-trained HTR source model, $\mathcal{M}_{S}$, it is necessary to seek information pertaining to the source data distribution within it. It is possible to gather statistics that approximate the source data distribution from normalization layers, such as batch normalization (BN). BN normalizes the outputs of a network layer on the basis of the statistics of a training mini-batch, thus ensuring that they have zero-mean and unit-variance. 

Formally, let $b \in \mathbb{R}^{B \times F}$ denote a mini-batch of $B$ feature vectors, each of size $F$. BN is defined as follows:
\begin{equation}
\label{eq:BN}
    \text{BN}(b) = \gamma \cdot \frac{b - \mu_{b}}{\sqrt{\sigma_{b}^2 + \epsilon}} + \beta
\end{equation}
\noindent Here, $\mu_b \in \mathbb{R}^{F}$ and $\sigma^{2}_b \in \mathbb{R}^{F}$ represent the mean and variance vectors, respectively, while $\epsilon$ is a small positive constant introduced for numerical stability. After applying standardization, the resulting mini-batch has zero-mean and unit-variance. However, this choice of unit variance is arbitrary. In practice, we allow the deep neural network to determine the distribution that best suits its requirements by incorporating element-wise scaling $\gamma \in \mathbb{R}^{F}$ and shifting $\beta \in \mathbb{R}^{F}$ parameters. Both $\gamma$ and $\beta$ are parameters that the model learns during training.

During the inference phase, it is essential to consider that BN might not be able to compute the $\mu_b$ and $\sigma^{2}_b$ statistics because the input may not constitute a mini-batch. To address this, BN stores exponentially weighted averages of the mean and variance vectors during the training phase, represented as $\mu_{\text{BN}}$ and $\sigma^{2}_{\text{BN}}$ respectively, and uses them during inference to perform standardization.

It is, therefore, possible to employ BN statistics in order to approximate the inaccessible source data feature distribution as a Gaussian distribution, $\mathcal{N}_S(\mu_{S}, \sigma^{2}_{S})$, where the $\mu_{S}$ and $\sigma^{2}_{S}$ are, respectively, the $\mu_{\text{BN}}$ and $\sigma^{2}_{\text{BN}}$ stored as explained above. Moreover, the violation of the i.i.d. assumption makes it necessary to perform cross-domain adaptation so as to reduce distribution discrepancy between source and target domains, thus ensuring that the model is ``usable'' for the target data. This is achieved by fine-tuning the layers that precede the BN layer in an unsupervised manner by forcing their extracted features to have mean and variance vectors similar to those of the source data.

Specifically, when given a mini-batch of target feature-vectors extracted before the BN layer, $b_T$, we compute its mean $\mu_{b_T}$ and variance $\sigma^{2}_{b_T}$ vectors, respectively. The target batch feature distribution is subsequently also approximated as $\mathcal{N}_{b_T}(\mu_{b_T}, \sigma^{2}_{b_T})$. We then employ the feature-averaged Kullback-Leibler (KL) divergence in order to align the target batch feature distribution with the approximated source feature distribution:

\begin{equation}
{\small
\begin{split}
\label{eq:align}
    & \mathcal{L}_a =  \mathcal{D}_{\text{KL}}\left(\mathcal{N}_{b_T} ||  \mathcal{N}_S \right) =\\
     &= \frac{1}{F}\sum_{i=1}^F \left ( \log \frac{\sqrt{\sigma^2_{S(i)}}}{\sqrt{\sigma^2_{b_T(i)}}} + \frac{\sigma^2_{b_T(i)} + (\mu_{b_T(i)} - \mu_{S(i)})^2}{2\sigma^2_{S(i)}}- \frac{1}{2} \right )
\end{split}
}
\end{equation}

Hereafter, we assume that the pre-trained HTR source model performs BN at least once. Note that this assumption is not particularly strong, as BN has become one of the most frequently used layers in deep neural networks~\cite{Bjorck:NIPS:2018, Santurkar:NIPS:2018}, including HTR architectures~\cite{Puigcerver:ICDAR:2017, puigcerver2018pylaia, Cascianelli:IJDAR:2022, Davoudi:NCA:2023, Vidal:PR:2023}. BN is typically applied to the output of convolutional layers, signifying that the operation is performed on a per-channel basis in all spatial locations. Furthermore, it must be noted that by fine-tuning only the layers preceding the BN layer, the remaining parts of the model, along with the normalization layer, can be made reusable for the target data. This implies a transfer of knowledge in order to solve the text recognition task.

Note that while the Align loss has been described in the context of a single BN layer, it can be applied to any or many of them simultaneously. In the latter case, an Align loss is calculated for each layer, with the final Align loss being the sum of those computed for the layers to which it is applied.

\subsubsection{Minimize.}
\label{subsubsec:minimize}

Merely aligning the feature distributions between the source and target domains does not ensure correct performance in the target domain, as it does not guarantee that the target features are sufficiently inherently discriminative to generate accurate predictions.

As detailed in Sec.~\ref{subsec:htr_model}, for a given input image, $x$, the recurrent block outputs a sequence of length $K$, where each time step, $y_{k}$, represents a probability distribution over $\Sigma'$. Ideally, these distributions should closely resemble a one-hot (delta) distribution, establishing a one-to-one relationship between each image frame and one character from the $\Sigma'$ alphabet. In order to encourage this in the SFUDA scenario, in which no labels are provided, we aim to minimize the uncertainty within each frame's distribution, i.e., its entropy. Specifically, given a mini-batch of $B$ sequences, each consisting of $K$ frames, the Minimize loss term is as follows:

\begin{equation}
\label{eq:minimize}
    \mathcal{L}_m = \frac{-1}{B \cdot K \cdot |\Sigma'_S|} \sum_{i=1}^{B} \sum_{k=1}^K \sum_{\pi \in \Sigma'_S} \left ( y_{x_{i_k}}^{\pi_k} \log y_{x_{i_k}}^{\pi_k} \right )
\end{equation}

Note that the $\mathcal{L}_m$ term is minimized when all frame-wise predictions are one-hot vectors, i.e., their probability distribution over $\Sigma'_S$ has zero entropy.

\subsubsection{Diversify.}
\label{subsubsec:diversify}

If we consider only the Align and Minimize term losses, a trivial solution might be to always produce the same frame vector. This would result in an informational collapse. In order to avoid such a trivial solution, it is necessary to introduce a third loss term that promotes diversity for model target predictions. This additional loss term, known as Diversify, encourages a uniform distribution of characters at an inter-frame level. The goal is to maximize the entropy within the batch-wise average per-frame probability distribution over $\Sigma'$ so as to obtain a class-balanced prediction. Mathematically, this can be expressed as follows:
\begin{equation}
\label{eq:diversify}
    \mathcal{L}_d = \frac{1}{K \cdot |\Sigma'_S|}  \sum_{k=1}^K \sum_{\pi \in \Sigma'_S} \frac{1}{B} \sum_{i=1}^{B} \left( y_{x_{i_k}}^{\pi_k} \log y_{x_{i_k}}^{\pi_k} \right)
\end{equation}

Note that the $\mathcal{L}_d$ term is minimized when the batch-wise average per-frame $\Sigma'_S$ distribution is class-balanced: this uniform distribution has the maximum entropy value. It is worth emphasizing that this diversity-promoting objective is intended to operate within a batch of data and not within a sequence. This distinction is essential in order to align with the CTC policy. For instance, let us consider an image containing only the letter ``a''. In this case, a valid output sequence, assuming a setting with $K=4$, could be ``aaaa''. Attempting to enforce a uniform character distribution throughout a sequence would conflict with the validity of such sequences.

Interestingly, minimizing entropy in individual predictions with the Minimize loss and maximizing entropy across the batch with the Diversify loss can be understood as maximizing the empirical mutual information with the true character sequence conditioned to the input features: the model becomes more certain about its predictions for each frame, while also ensuring that, for many predictions, it does not become biased or uninformative by always predicting the same character.

\subsubsection{Final loss.}
\label{subsubsec:final_loss}

The overall loss function is a weighted average of the Align, Minimize and Diversify terms:

\begin{equation}
\label{eq:amd_loss}
    \mathcal{L} = w_a \cdot \mathcal{L}_a + w_m \cdot \mathcal{L}_m  + w_d \cdot \mathcal{L}_d
\end{equation}
\noindent where $w_a$, $w_m$, and $w_d$ are hyper-parameters controlling the importance of each term in the loss. 

\subsection{Model selection}
\label{subsec:model_selection}

With the initial inability to ensure an inverse correlation between loss and performance in the target domain, we resort to employing a target-domain validation set as our selection method. The datasets considered comprised partitions for training, validation and testing. We utilized the training partition to fine-tune the source model via AMD, ultimately choosing the model that maximized performance on the validation set. Finally, the result presented is the performance of this selected model on the test partition. 

We are aware that the proposed model selection strategy is not realistic; however, model selection is currently an open problem within the SFUDA field. Moreover, this challenge extends to adjacent domains such as Domain Generalization (DG). In~\cite{Gulrajani:ICLR:2021}, Gulrajani and Lopez-Paz emphasize that ``\emph{a DG algorithm without a model selection strategy remains incomplete}'' and propose several model selection strategies. However, given that we \emph{do} have knowledge of the target domain but lack the corresponding labels, none of the proposed strategies can be considered in the present work. We have included supplementary material that illustrates the correlation between the loss terms and the performance in the target domain (see ``Supplementary Material''). This additional information is included with the objective of providing further insights into the issue of model selection.

\section{Experimental set up}
\label{sec:set_up}

This section covers the experimental setup, including corpora, evaluation protocol and implementation details.

\subsection{Datasets}

\subsubsection{Real data.}
\label{subsubsec:real_ds}

\begin{figure}[!ht]
    \centering
    \includegraphics[width=0.7\textwidth]{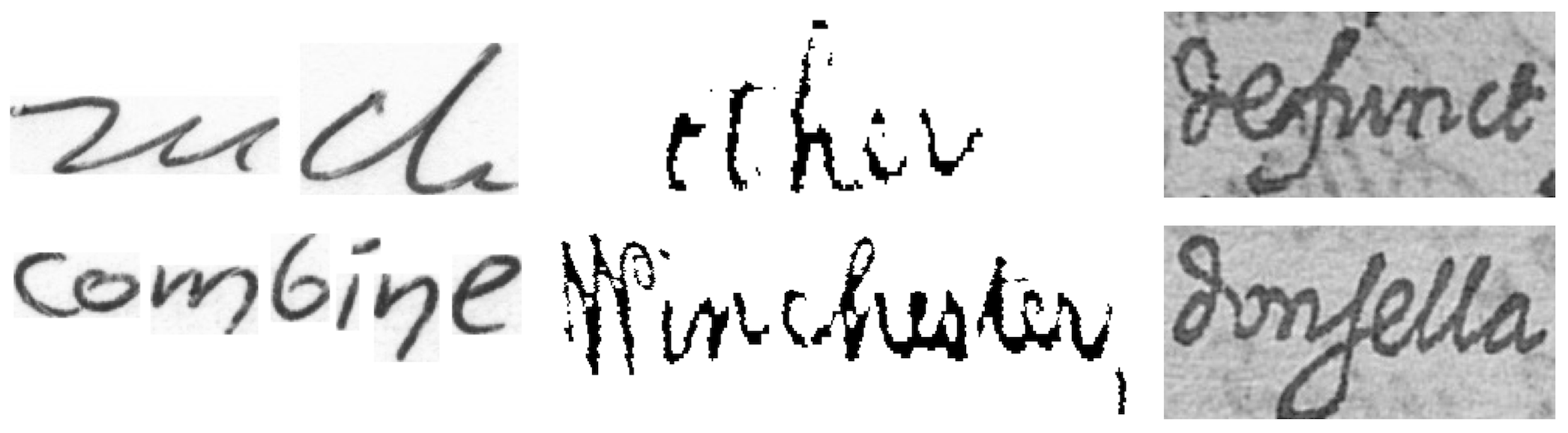}
    \caption{Word samples of the three real corpora used in the experiments. From left to right: IAM, GW and ESPOSALLES.}
    \label{fig:real_data}
\end{figure}

We extensively assessed the performance of our proposal through the use of three distinct publicly available datasets with different particularities: IAM~\cite{Marti:IJDAR:2002}, GW~\cite{Fischer:PRL:2012} and ESPOSALLES~\cite{Romero:PR:2013}. These datasets encompass multiple or single writers and comprise documents from both historic and modern periods, written in English or Catalan. Table~\ref{tab:corpora} provides a summary of the characteristics of these corpora, while Fig.~\ref{fig:real_data} shows some excerpts from each dataset.

We assessed the HTR task at a word recognition level using the standard training, validation and test splits provided by each dataset. It is important to note that ESPOSALLES provides only training and test sets, and we, therefore, allocated 10\% of the training set for validation.

In order to comprehensively assess our proposal in terms of style variability, size and alphabet, we propose the use of two evaluation scenarios when employing real data. In the first scenario, denoted as the \textbf{Single-source scenario}, we train the source model using only one of the datasets considered and then individually adapt it to the remaining ones. In the second scenario, referred to as the \textbf{Multi-source scenario}, we follow a leave-out-one approach: we train the source model using two of the three datasets considered and adapt it to the ``one-out'' dataset.\footnote{We ensure that each training batch is dataset-balanced---an equal number of samples from each dataset---in order to prevent bias towards any specific collection.}

\begin{table}[!ht]
\caption{Overview of the corpora used in this work, depicting their characteristics (number of samples, size of the alphabet, number of writers, time period and language).}
\label{tab:corpora}
\centering
\setlength{\tabcolsep}{6pt}
\renewcommand{\arraystretch}{0.9}
\begin{tabular}{l c c c c c}
\toprule[1pt]
 & \textbf{Images} & \textbf{Alphabet}& \textbf{Writers} & \textbf{Period} & \textbf{Language} \\
\cmidrule(lr){2-6} 
\textbf{IAM} & 115\,320 & 78& 657 & Modern & English \\
\textbf{GW} & 4\,894 & 67& 1 & Historic & English \\
\textbf{ESPOSALLES} & 39\,527 & 60& 1 & Historic & Catalan \\
\bottomrule[1pt]
\end{tabular}
\end{table}

\subsubsection{Synthetic data.}
\label{subsubsec:synth_ds}

Real-world HTR datasets are replete with nuances. While HTR often achieves impressive performances, these very nuances pose challenges in OOD settings, often resulting in a performance drop. This prompted us to question whether starting with a generic controlled synthetic dataset could lead to better adaptation to real-world data. It is important to note that the best-performing model within an in-domain context may not necessarily be the best choice for an OOD scenario. This led us to favor the idea of initially having a model that performs very well in OOD settings, given the fact that we know it will later be adapted to specific domains.

\begin{figure}
    \centering
    \includegraphics[width=0.8\textwidth]{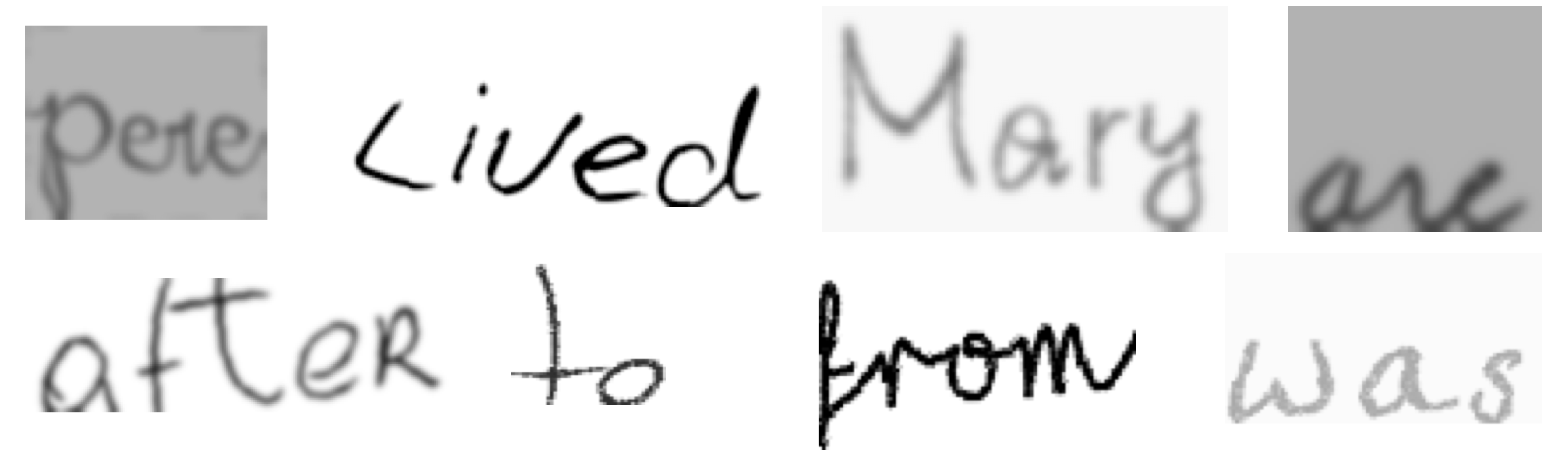}
    \caption{Word samples produced by the synthetic generator used in this work.}
    \label{fig:synthetic_data}
\end{figure}

There are two primary approaches for synthetic data generation for HTR: one utilizes TrueType fonts~\cite{Ingle:ICDAR:2019, Krishnan:ECCV:2016, Kang:WACV:2020}, while the other employs generative models~\cite{Bhunia:CVPR:2019, Kang:TPAMI:2021, Pippi:CVPR:2023}. While the latter is a promising field, it still cannot guarantee 100\% accuracy as regards generating correct words owing to its recent emergence. Additionally, it often uses datasets such as IAM for training, which could lead to misleading results here. We, therefore, opted to use the exact synthetic pipeline generation based on TrueType fonts described in~\cite{Krishnan:ECCV:2016, Kang:WACV:2020}. Some samples of these synthetic words are shown in Fig.~\ref{fig:synthetic_data}.

\subsection{Evaluation metrics}
\label{subsec:eval_protocal}

The Character Error Rate (CER) and Word Error Rate (WER) were used in order to measure model performance. CER is defined as the normalized Levenshtein distance~\cite{Levenshtein:SPD:1966} between ground-truth and prediction while WER is the percentage of improperly recognized words.

\subsection{Implementation details}
\label{subsec:implementation_details}
The CRNN scheme is based on that used typically for HTR~\cite{Puigcerver:ICDAR:2017, puigcerver2018pylaia, Cascianelli:IJDAR:2022, Davoudi:NCA:2023, Vidal:PR:2023}. Specifically, we use Puigcerver's configuration~\cite{Puigcerver:ICDAR:2017}, consisting of five convolutional blocks and five Bidirectional Long Short-Time Memory layers. These convolutional blocks already include BN layers, making them directly compatible with our approach; therefore, no additional layers are needed. 

The evaluation pipeline consisted of two stages: (i) training the source model, and (ii) adapting it to the target dataset using AMD. In both cases, we used the ADAM optimizer to train the models with a batch size of 16 elements. The learning rate was fixed at $3\cdot10^{-4}$ in (i), whereas in (ii), we conducted a random search between $3\cdot10^{-4}$ and $10^{-5}$, as we were fine-tuning an already trained model. We stopped the training using an early stopping strategy with a patience of 20 epochs, retaining the weights that minimize the CER metric in the validation partition. This aligns with the model selection strategy for SFUDA discussed in Sec.~\ref{subsec:model_selection}.

With regard to data preprocessing, each word image is resized to a height of 128 pixels, maintaining the aspect ratio (signifying that each sample might differ in width) and converting them to grayscale, with no further preprocessing. During the training of source models, we included a data augmentation step, which was applied online within the data loader. This ensured that each batch was randomly augmented in order to enhance training. For more specific information regarding the data augmentation pipeline, we refer the reader to the ``Supplementary Material''.

\section{Results}
\label{sec:results}

We conducted a random search over a range of values for the AMD weight term losses, specifically exploring values in the set $[0, 1, 5, 10, 25, 50]$. Further details on the impact of each hyper-parameter can be found in Sec.~\ref{subsec:ablation}. Tables~\ref{tab:results_all} and~\ref{tab:results_comp} present the results obtained for the proposed experimental scheme in terms of the CER and WER metrics. For the sake of clarity, the actual values of the weight term losses, along with the BN layers employed for the adaptation, are not reported here but can be found in the ``Supplementary Material''.

\subsection{Performance analysis}
\label{subsec:perf_results}

An initial observation is that the proposed AMD method consistently outperforms the baseline (the source model directly evaluated over the target domain), regardless of the source-target configuration. It is, therefore, logical to conclude that it is always better to attempt adaptation, even in those cases in which there is limited room. Nevertheless, the extent of improvement depends on the specific source-target configuration. We shall now analyze each of the scenarios considered in detail.

\begin{table}[ht!]
    \caption{Results in terms of the CER (\%) metric. Baseline refers to the source model directly evaluated over the target domain. Boldface is used to highlight the lowest error achieved by the AMD method for a specific target dataset. (\textbf{\textcolor{deepgreen}{X\%}}) depicts the reduction rate in the CER metric achieved after AMD adaptation.}
    \label{tab:results_all}
    \centering
    \renewcommand{\arraystretch}{.9}
    \resizebox{1\textwidth}{!}{
    \begin{tabular}{lcccc}
    \toprule[1pt]
                   &                                    & \multicolumn{3}{c}{Target corpus}                                                                                                                                                                                                                         \\ \cmidrule(lr){3-5} 
                   &                                    & IAM                                                                           & GW                                                                            & ESPOSALLES                                                                                \\ \cmidrule(lr){3-5} 
                   &                                    & Baseline $\rightarrow$ AMD                                                    & Baseline $\rightarrow$ AMD                                                    & Baseline $\rightarrow$ AMD                                                                \\ \cmidrule(lr){1-5} 
    \textit{Single source}                              & \multicolumn{4}{l}{}                                                                                                                                                                                                                                      \\
                   & IAM                                & -                                                                             & 55.5 $\rightarrow$ 20.6 (\textbf{\textcolor{deepgreen}{62.9\%}})              & \phantom{0}52.9 $\rightarrow$ 30.7 (\textbf{\textcolor{deepgreen}{41.9\%}})               \\
                   & GW                                 & 98.1 $\rightarrow$ 95.5 (\textbf{\textcolor{deepgreen}{2.6\%}})               & -                                                                             & 109.7 $\rightarrow$ 86.1 (\textbf{\textcolor{deepgreen}{21.5\%}})                         \\
                   & ESPOSALLES                         & 79.9 $\rightarrow$ 78.7 (\textbf{\textcolor{deepgreen}{1.4\%}})               & 78.7 $\rightarrow$ 68.9 (\textbf{\textcolor{deepgreen}{12.4\%}})              & -                                                                                         \\ \cmidrule(lr){1-5} 
    \textit{Multi-source}                               & \multicolumn{4}{c}{}                                                                                                                                                                                                                                      \\
                   & IAM-GW                             & -                                                                             & -                                                                             & \phantom{0}48.1 $\rightarrow$ 35.9 (\textbf{\textcolor{deepgreen}{25.4\%}})               \\
                   & IAM-ESPOSALLES                     & -                                                                             & 51.7 $\rightarrow$ 20.7 (\textbf{\textcolor{deepgreen}{59.9\%}})              & -                                                                                         \\
                   & GW-ESPOSALLES                      & 33.6 $\rightarrow$ 33.1 (\textbf{\textcolor{deepgreen}{1.4\%}})               & -                                                                             & -                                                                                         \\ \cmidrule(lr){1-5}
    \textit{Synthetic data}                             & \multicolumn{4}{c}{}                                                                                                                                                                                                                                      \\
                   &                                    & 25.5 $\rightarrow$ \textbf{17.8} (\textbf{\textcolor{deepgreen}{30.2\%}})     & 26.0 $\rightarrow$ \textbf{13.8} (\textbf{\textcolor{deepgreen}{46.9\%}})     & \phantom{0}29.7 $\rightarrow$ \textbf{12.7} (\textbf{\textcolor{deepgreen}{57.2\%}})      \\ \bottomrule[1pt]
    \end{tabular}
    }
\end{table}

\paragraph{\textbf{Single-source scenario.}} Using the IAM dataset as the data source consistently leads to the highest improvement rates. This can almost certainly be attributed to the fact that the pre-trained HTR models start with lower initial error rates. Furthermore, a substantial overlap with the target alphabet ($\geq 94\%$) further boosts these adaptations. To illustrate, targeting the GW dataset is a scenario particularly favored by this alphabet overlap. In this specific case, we achieve the highest relative improvement (63\%) of all single-source scenarios, reducing the CER metric from 55.5 to 20.6. In contrast, utilizing the GW dataset as the source dataset entails adapting to initial CER values of over 90\%. This means that, regardless of the extent of improvement (up to 23 points in some cases), the method is not able to recuperate a pre-trained model that initially performed so poorly. A similar situation arises when the ESPOSALLES dataset is employed as source, which starts with an average CER of 80\%.

\paragraph{\textbf{Multi-source scenario.}} These cases have better baselines, which means that using multiple datasets to pre-train the HTR model helps recognize OOD samples to some extent. However, this does not always directly correlate with the outcome of the adaptation. We can identify three representative situations: (i) when IAM is the target dataset, employing multi-source data \emph{does} assist in adaptation, as it results in a 1-point increase in the percentage of relative improvement; (ii) in the case of GW as the target dataset, incorporating multi-source data does not provide meaningful additional value, as the performance in the target domain after adaptation closely resembles the results achieved in the best single-source scenario, and (iii) when targeting ESPOSALLES using a multi-source model, it actually degrades the performance when compared to the best single-source adapted model, despite starting from a lower CER value.

\paragraph{\textbf{Synthetic data scenario.}} In this scenario, the baseline consistently outperforms the real-to-real case. This is because the model trained with synthetic data is less adjusted to the specific nuances of each dataset, making it inherently more general. Additionally, synthetic data allows for a 100\% overlap with any target alphabet, contributing to its effectiveness. Nevertheless, even when reducing the margin for adaptation, AMD outperforms the the baseline results in all cases. We can, therefore, conclude that the use of synthetic data for SFUDA with AMD clearly improves the performance in OOD samples, while being more cost-effective and more prone to adaptation than real data.

\paragraph{\textbf{Further insights.}} In certain scenarios (e.g., GW as the source dataset and ESPOSALLES as the target), achieving an improvement of approximately 80\% is necessary in order to create a functional model. This is not merely a matter of adaptation, but rather signifies that no useful knowledge has been gained for OOD scenarios. However, outside these specific situations, we have consistently observed that the decrease in error rates is less pronounced in cases involving the IAM dataset. One particularity of this dataset is that it is actually a meta-dataset, essentially representing scenarios in which there are multiple different writers. Some additional experiments (see ``Supplementary Material'') seemed to suggest that this may be a potential reason for the failure to adapt, which is discussed in greater detail in the supplementary material.

\subsection{Comparison with the state of the art}
\label{subsec:comparison}

Table~\ref{tab:results_comp} compares our results to state-of-the-art DA methods, providing context for the significance of our achieved outcomes. Within the existing literature, these approaches are evaluated under two distinct settings: (i) treating IAM training and test splits as separate domains (given the multi-writer nature of this collection), and (ii) utilizing synthetic data as the source domain.\footnote{Pre-adaptation performance is equal in all cases in order to ensure fair comparison and mitigate any performance disparities arising from pre-trained model variations.} In the former setting, all methods show similar performance due to limited domain distribution shift---adaptation has, therefore, little room for improvement. Nevertheless, our AMD approach demonstrates superior performance, particularly in CER. In the latter scenario, we compare our results, already depicted in Table~\ref{tab:results_all}, with those reported by Kang et al. Our findings reveal notable improvements over Kang et al.'s results, especially in ESPOSALLES (with a nearly 50\% reduction in CER) and GW datasets, albeit trailing behind in the IAM dataset. This discrepancy in IAM's performance can be attributed to its meta-dataset nature, as previously discussed. Notably, while our CER values showcase improvement, our WER values do not exhibit similar advancements. This underscores the primary limitation of AMD: the absence of language modeling adaptation, which warrants further exploration. In summary, we can state that our source-free approach, AMD, is competitive against traditional DA methods that require source data for adaptation. 

\begin{table}[ht!]
    \caption{Results in terms of the CER / WER (\%) metrics. Boldface is used to highlight the lowest error achieved for a specific target dataset.}
    \label{tab:results_comp}
    \centering
    \setlength{\tabcolsep}{4pt}
    \renewcommand{\arraystretch}{1.2}
    \resizebox{1\textwidth}{!}{
    \begin{tabular}{lccccccc}
    \toprule[1pt]
                                                       & \multirow{2}{*}{Source-Free?} & & \textit{Source}     & IAM$_{\text{Train}}$       & Synthetic                         & Synthetic                     & Synthetic                         \\  
                                                       &                               & & \textit{Target}     & IAM$_{\text{Test}}$        & IAM                               & GW                            & ESPOSALLES                        \\ \cmidrule(lr){1-8} 
    Zhang et al. [CVPR19]~\cite{Zhang:CVPR:2019}       & $\times$                      & &                     & 8.5 / 22.2                 & -                                 & -                             & -                                 \\
    Tang et al. [ICASSP22]~\cite{Tang:ICASSP:2022}     & $\times$                      & &                     & 7.2 / \textbf{15.5}        & -                                 & -                             & -                                 \\
    Kang et al. [WACV20]~\cite{Kang:WACV:2020}         & $\times$                      & &                     & 6.8 / 17.3                 & \textbf{14.1} / \textbf{34.9}     & 16.3 / \textbf{40.0}          & 21.0 / 50.0                       \\ \cdashlinelr{1-8}
    AMD                                                & \checkmark                    & &                     & \textbf{5.9} / 17.5        & 17.8 / 49.8                       & \textbf{13.8} / 44.8          & \textbf{12.7} / \textbf{44.5}     \\ \bottomrule[1pt]
    \end{tabular}
    }
\end{table}

\subsection{Ablation study with the loss terms}
\label{subsec:ablation}
In order to assess the particular impact of each term of the AMD loss, Table~\ref{tab:loss_comb} presents the median decrease in the CER metric of this ablation study. The results indicate that while a significant improvement is observed with the inclusion of only the Minimize ($\mathcal{L}_m$) loss term combined with either Align ($\mathcal{L}_a$) or Diversify ($\mathcal{L}_d$), it becomes evident that using all three loss terms is the most effective approach. 

\begin{table}[ht!]
\caption{Impact of each of the three regularization terms introduced in AMD---Align, $\mathcal{L}_a$; Minimize, $\mathcal{L}_m$; and, Diversify, $\mathcal{L}_d$. We report the median decrease in CER over the aggregation of the Top-1 result in each source-target configuration for a better notion of their impact.}
\label{tab:loss_comb}
\centering
\setlength{\tabcolsep}{4pt}
\renewcommand{\arraystretch}{0.9}
\begin{tabular}{cccccccc}
\toprule[1pt]
$\mathcal{L}_a$         & \checkmark    & -             & -             & \checkmark    & \checkmark    & -             & \checkmark    \\
$\mathcal{L}_m$         & -             & \checkmark    & -             & \checkmark    & -             & \checkmark    & \checkmark    \\
$\mathcal{L}_d$         & -             & -             & \checkmark    & -             & \checkmark    & \checkmark    & \checkmark    \\ \cmidrule(lr){1-8} 
Median decrease (\%)    & 4.5           & 7.2           & -4.1          & 10.3          & 1.5           & 9.1           & \textbf{13.1} \\ \bottomrule[1pt]
\end{tabular}
\end{table}

\section{Conclusions}
\label{sec:conclusions}

We have introduced a new SFUDA approach for HTR: the Align, Minimize and Diversify (AMD) method. We overcome the need to access source data by integrating three different regularization terms. The Align term reduces the feature distribution discrepancy between source and target data by aligning BN statistics. The Minimize term drives the model towards confident predictions by forcing the outputs to resemble one-hot distributions. Finally, the Diversify term prevents uniform outputs by promoting diverse sequences throughout the target data. Extensive experimentation with three HTR benchmarks and synthetic data demonstrated that the AMD method is competitive against methods incorporating source data during adaptation, as it notably improves the performance in OOD samples, particularly when using synthetic data. 

\paragraph{\textbf{Limitations and future work.}} SFUDA in HTR requires two levels of alignment: graphical and language modeling. AMD addresses the first, and while successful, the results emphasize the importance of addressing both for improved adaptation performance. It is necessary to carry out more work in this direction. Furthermore, we observed that having a multi-writer target dataset adds an additional layer of complexity. Future adaptations must, therefore, also have the ability to handle multi-target (multi-writer) data effectively. Lastly, we acknowledge that relying on BN for the Align loss somewhat limits the applicability of AMD. Overcoming this constraint presents a promising opportunity to enhance the method's versatility and eliminate architectural dependencies.

\newpage

\begin{appendix}

\centering\section*{SUPPLEMENTARY MATERIAL}

\noindent We provide additional details and results in this supplementary material, which is organized as follows:
\begin{itemize}
    \item Section~\ref{sec:amd_pseudocode}: AMD pseudocode.
    \item Section~\ref{sec:data_aug}: data augmentation.
    \begin{itemize}
        \item Section~\ref{subsec:train_aug}: data augmentation for pre-training source models.
        \item Section~\ref{subsec:synth_aug}: synthetic word generator augmentations.
    \end{itemize}
    \item Section~\ref{sec:vocab_overlap}: overlap of alphabet for the considered datasets.
    \item Section~\ref{sec:corr_loss_perf}: insights on model selection based on performance and AMD loss correlation.
    \item Section~\ref{sec:hyper_results}: hyper-parameter configuration of the results presented.
    \item Section~\ref{sec:writer_adap}: insights on multi-target (multi-writer) AMD adaptation.
    \item Section~\ref{sec:bn_ablation}: ablation study with the BN layers.
\end{itemize}

\newpage

\section{AMD pseudocode}
\label{sec:amd_pseudocode}

Below, we present pseudocode for the implementation of the Align, Minimize and Diversify (AMD) method in PyTorch. We describe the AMD adaptation when only the statistics from one BN layer are considered in order to reduce the distribution discrepancy between source and target data. If more BN layers were to be considered, we would obtain one Align loss per layer, with the final Align loss being the sum of all of them.

\begin{lstlisting}[style=mystyle, caption={AMD PyTorch pseudocode.}, label=amd_pseudocode]
# f: pre-trained source model
# bn_mean, bn_var: statistics of the BN layer of f
# wa, wm, wd: coefficients of the align, minimize and
# diversify losses

def entropy(x, eps=1e-4):
    x = x.softmax(dim=1)
    # Numerical stability (avoid zero probabilities):
    x = torch.clamp(x, min=eps, max=1.0)
    h = -1 * ((x * torch.log(x)).sum(dim=1))
    h = h.mean()
    return h

for x in loader:
    # Compute representations:
    # 1) Model output: z
    # 2) BN stats: x_mean, x_var
    z, x_mean, x_var = f(x)

    # Align loss
    a_loss_1 = torch.sqrt(bn_var) / torch.sqrt(x_var)
    a_loss_1 = torch.log(a_loss_1)
    a_loss_2 = x_var + (x_mean - bn_mean).pow(2)
    a_loss_2 /= 2 * bn_var
    a_loss = a_loss_1 + a_loss_2 - 0.5
    a_loss = a_loss.mean()

    # Minimize loss
    B, T, F = z.shape
    m_loss = entropy(z.reshape(B * T, F))

    # Diversify loss
    d_loss = entropy(z.mean(dim=0))

    # AMD loss
    loss = wa * a_loss + wm * m_loss - wd * d_loss

    # Optimization step
    loss.backward()
    optimizer.step()
\end{lstlisting}

\section{Data augmentation}
\label{sec:data_aug}

We consider two different augmentation pipelines: one with which to augment training real data samples in the pre-training stage and the other with which to generate realistic synthetic word images.

\subsection{Pre-training augmentations}
\label{subsec:train_aug}

The following operations are sequentially performed in order to generate augmented training samples:
\begin{itemize}
    \item Elastic deformation is applied with a displacement magnitude of 20 and a probability of 20\%.
    \item A random rotation of up to 3 degrees is applied with a 50\% probability.
    \item Color jittering, including adjustments to brightness, contrast, saturation and hue, is performed with a probability of 20\%. {\small\texttt{ColorJitter(brightness=0.4, contrast=0.4, saturation=0.2, hue=0.1)}} in PyTorch.
    \item Gaussian blur is applied with a probability of 20\% and a kernel size of 23.
\end{itemize}

\subsection{Synthetic word generator augmentations}
\label{subsec:synth_aug}

The following operations are randomly applied to synthetic word images in order to create realistic deformations that can be found in handwritten data:
\begin{itemize}
    \item Elastic deformation is applied 20\% of the time, with a displacement magnitude of 20 and a smoothness factor of 4.
    \item Gaussian blur is applied with a 20\% probability and a kernel size of 23.
    \item Random patches are erased, causing pixel colors to change to the background color, with a 20\% probability. We randomly select a scale factor of between 0.01 and 0.03 of the input image size for the erased patches, along with a randomly chosen aspect ratio ranging from 0.2 to 3.2 for these erased patches.
    \item A random perspective transformation with a distortion scale factor of 0.2 is applied with a 20\% probability.
    \item Adjustments to brightness, contrast, saturation and hue, as described in~\cite{Liu:ECCV:2016}, are applied with a 50\% probability. {\small\texttt{RandomPhotometricDistort(p=0.5)}} in PyTorch.
    \item A random affine transformation, maintaining center invariant, is applied with a 50\% probability. {\small\texttt{RandomAffine(degrees=5, translate=(0.05, 0.05), scale=(0.95, 1.05), shear=(-5, 5, -1.5, 1.5))}} in PyTorch.
\end{itemize}

\section{Overlap of alphabet}
\label{sec:vocab_overlap}

The real-world datasets considered do not share an alphabet. Table~\ref{tab:vocab_overlap} shows the rounded-up degree of overlap of each source dataset alphabet with respect to each target dataset alphabet.

\begin{table}[ht!]
\setlength\tabcolsep{6pt}
\caption{Rounded-up degree of overlap (\%) of each source dataset alphabet with respect to each target dataset alphabet.}
\label{tab:vocab_overlap}
\centering
\renewcommand{\arraystretch}{1.2}
\begin{tabular}{lcccc}
\toprule[1pt]
 &                                                  & \multicolumn{3}{c}{Target}\\ \cmidrule(lr){3-5}
 &                                                  & IAM   & GW   & ESPOSALLES \\ \cmidrule(lr){1-5} 
\multirow{3}{*}{\rotatebox{90}{Source}} & IAM       & -     & 100  & 97         \\ 
                                        & GW        & 86    &  -   & 92         \\ 
                                        & ESPOSALLES& 74    & 85   & -          \\ \bottomrule[1pt]
\end{tabular}
\end{table}

\newpage

\section{More on model selection}
\label{sec:corr_loss_perf}

\begin{figure}[!ht]
    \centering
    \includegraphics[width=1\textwidth]{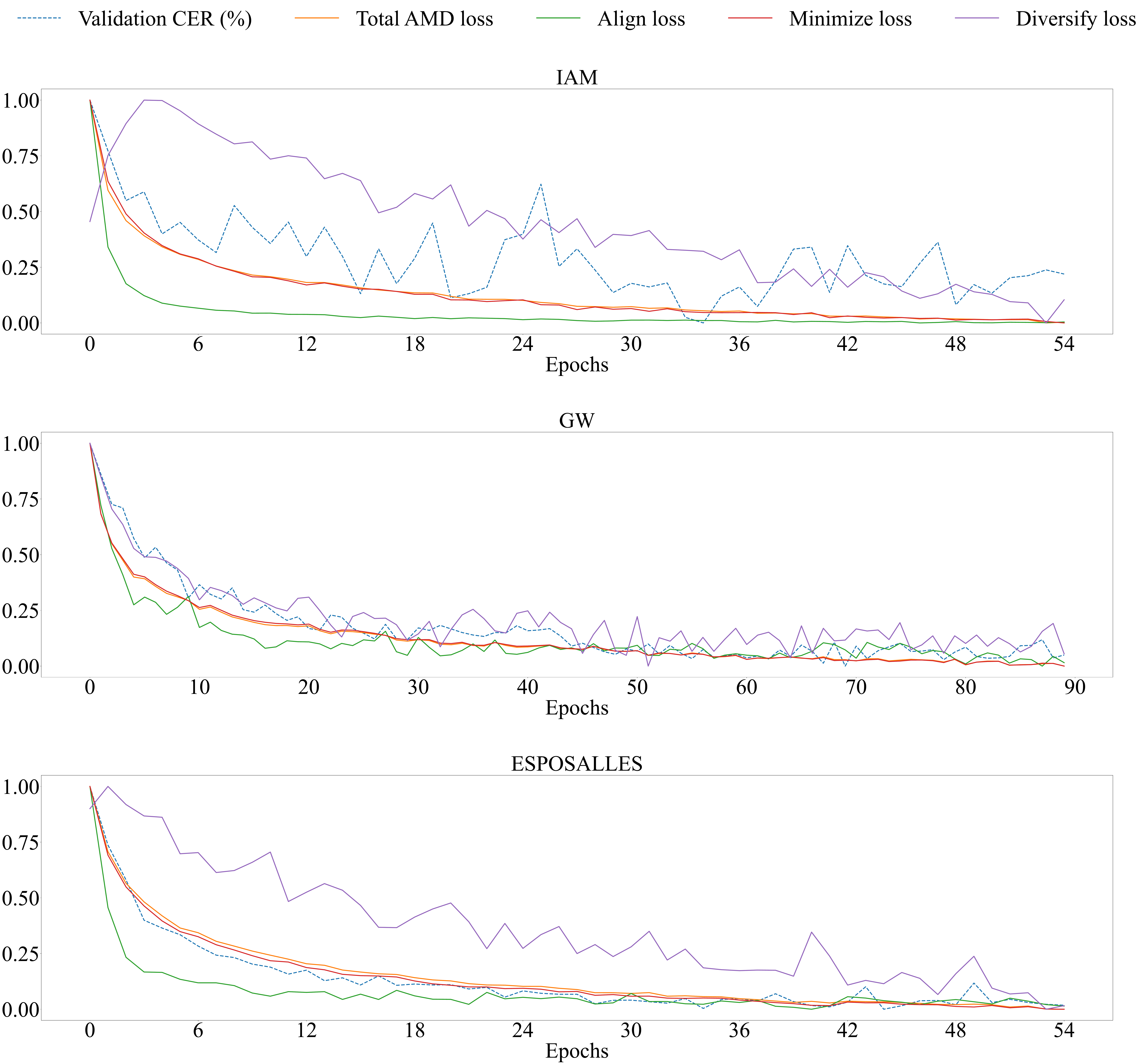}
    \caption{Normalized progression of the validation CER metric, the AMD loss and each individual regularization term  across epochs for each source-target configuration. In all cases, synthetic data is used as the source data.}
    \label{fig:loss_corr}
\end{figure}

Model selection remains an open issue within the (SFU)DA field and closely related areas such as DG. This section aims to provide additional insights on this topic by examining whether the AMD loss can serve as a proxy for CER in the absence of labeled target domain data. Accordingly, Fig.~\ref{fig:loss_corr} illustrates the normalized progression of the validation CER metric, the AMD loss and each individual regularization term across epochs. We use synthetic data for pre-training the HTR model and report the evolution in the individual AMD adaptation to each of the three real datasets considered. We report these scenarios not only for the sake of clarity but also because they correspond to the best overall results (as seen in Table \textcolor{red}{2}) and exemplify the varied correlations between performance and loss observed in other scenarios.

Examining Fig.~\ref{fig:loss_corr}, we observe instances where there exists a discernible correlation between the AMD loss and the CER, as exemplified by the GW and ESPOSALLES scenarios. However, this relationship is not consistent across different scenarios. This highlights the need for further exploration into the dynamics of model selection within the SFUDA domain.

\newpage

\section{Hyper-parameters of the results presented}
\label{sec:hyper_results}

Table~\ref{tab:hyper_results} shows the weight term losses, the BN layers and the fixed learning rate considered during the AMD adaptation of the results reported in Table \textcolor{red}{2}.

\begin{table}[ht!]
    \caption{Hyper-parameter values corresponding to the results presented in Table \textcolor{red}{2} for each source-target configuration. In this context, $lr$ represents the fixed learning rate throughout the entire adaptation process, while $w_a$, $w_m$, and $w_d$ control the importance of each term in the AMD loss function. Additionally, \emph{BN} refers to the layer or layers considered so as to align the source and target distributions. The CRNN architecture employed in this work includes four BN layers. A numerical identifier is used in order to distinguish the BN layer, starting from 0.}
    \label{tab:hyper_results}
    \centering
    \resizebox{1\textwidth}{!}{
    \begin{tabular}{lcccc}
    \toprule[1pt]
                   &                      & \multicolumn{3}{c}{Target corpus}\\ \cmidrule(lr){3-5} 
                   &                      & IAM                  & GW                   & ESPOSALLES\\ \cmidrule(lr){1-5} 
    \textit{Single-Source}  & \multicolumn{1}{l}{} & \multicolumn{1}{l}{} & \multicolumn{1}{l}{} & \multicolumn{1}{l}{}\\
                   & IAM                  & -                                                                                                               & \begin{tabular}[c]{@{}c@{}}$l_r=3\cdot10^{-4}$\\ $w_a=25, w_m=50, w_d=5$\\$\text{BN}=[2, 3, 4]$\end{tabular}     & \begin{tabular}[c]{@{}c@{}}$l_r=3\cdot10^{-4}$\\ $w_a=25, w_m=50, w_d=1$\\$\text{BN}=[4]$\end{tabular}\\ \cmidrule(lr){2-5} 
                   & GW                   & \begin{tabular}[c]{@{}c@{}}$l_r=10^{-5}$\\ $w_a=50, w_m=10, w_d=10$\\$\text{BN}=[1, 2, 3, 4]$\end{tabular}     & -                                                                                                                 & \begin{tabular}[c]{@{}c@{}}$l_r=10^{-5}$\\ $w_a=50, w_m=10, w_d=25$\\$\text{BN}=[3]$\end{tabular}\\ \cmidrule(lr){2-5} 
                   & ESPOSALLES           & \begin{tabular}[c]{@{}c@{}}$l_r=10^{-5}$\\ $w_a=5, w_m=1, w_d=1$\\$\text{BN}=[1, 2, 4]$\end{tabular}           & \begin{tabular}[c]{@{}c@{}}$l_r=3\cdot10^{-4}$\\ $w_a=10, w_m=5, w_d=10$\\$\text{BN}=[2, 4]$\end{tabular}        & -\\ \cmidrule(lr){1-5}
    \textit{Multi-Source}   &                      &                      &                      &                                           \\  
                   & IAM-GW               & -                                                                                                               & -                                                                                                                 & \begin{tabular}[c]{@{}c@{}}$l_r=3\cdot10^{-4}$\\ $w_a=25, w_m=50, w_d=1$\\$\text{BN}=[4]$\end{tabular}\\ \cmidrule(lr){2-5} 
                   & IAM-ESPOSALLES       & -                                                                                                               &\begin{tabular}[c]{@{}c@{}}$l_r=3\cdot10^{-4}$\\ $w_a=50, w_m=10, w_d=0$\\$\text{BN}=[1, 3, 4]$\end{tabular}      & -\\ \cmidrule(lr){2-5} 
                   & GW-ESPOSALLES        & \begin{tabular}[c]{@{}c@{}}$l_r=3\cdot10^{-4}$\\ $w_a=25, w_m=50, w_d=5$\\$\text{BN}=[3, 4]$\end{tabular}      & -                                                                                                                 & - \\ \cmidrule(lr){1-5} 
    \textit{Synthetic Data} &                      &                      &                      &                                    \\
                   &                      & \begin{tabular}[c]{@{}c@{}}$l_r=3\cdot10^{-4}$\\ $w_a=25, w_m=10, w_d=5$\\$\text{BN}=[4]$\end{tabular}         & \begin{tabular}[c]{@{}c@{}}$l_r=3\cdot10^{-4}$\\ $w_a=1, w_m=1, w_d=1$\\$\text{BN}=[1, 2, 4]$\end{tabular}       & \begin{tabular}[c]{@{}c@{}}$l_r=3\cdot10^{-4}$\\ $w_a=50, w_m=50, w_d=25$\\$\text{BN}=[1, 4]$\end{tabular}\\ \bottomrule[1pt]
    \end{tabular}
    }
\end{table}

\newpage

\section{Writer adaptation}
\label{sec:writer_adap}

\begin{figure}[!ht]
    \centering
    \includegraphics[width=0.6\textwidth]{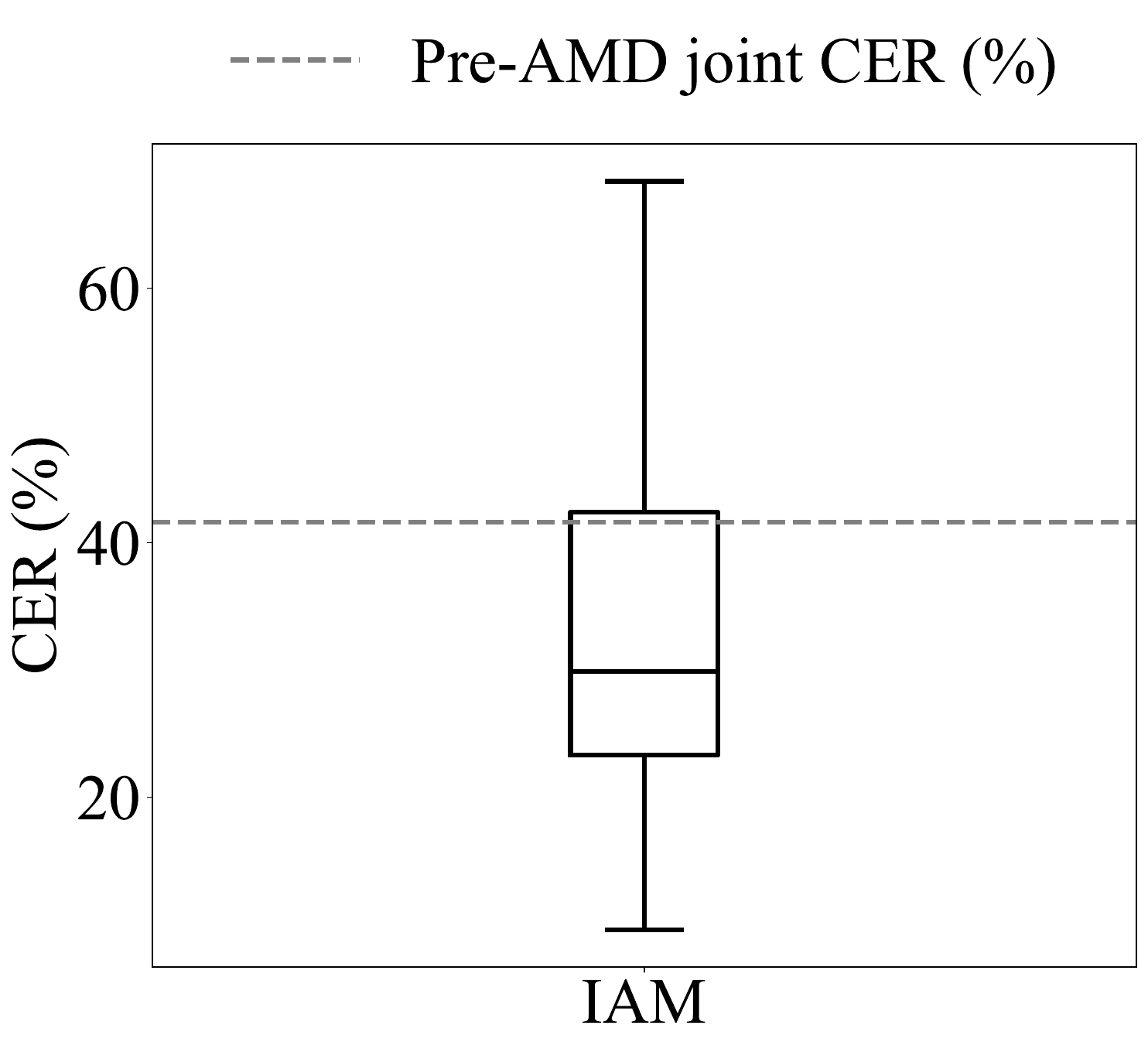}
    \caption{Distribution of CER values across the different writers of IAM after AMD adaptation. The dashed \emph{joint} line indicates the baseline (no AMD). We used synthetic data as the source data used for pre-training.}
    \label{fig:writers}
\end{figure}

The results from Table \textcolor{red}{2} showed a less pronounced decrease in the CER metric when targeting the IAM dataset. This observation suggests that the problem may not lie with the methodology or the pre-training of the model but with the target dataset itself. To delve deeper into this issue, we computed the CER for each individual writer within this dataset.\footnote{We chose synthetic data as the source data because it yields better results and also ensures a complete alphabet overlap with the target, thus preventing differences arising from alphabet discrepancies.} The wide ranging outcomes, illustrated in Fig.~\ref{fig:writers}, suggest that the IAM dataset should not be treated as a singular entity due to the disparate results across different writers. Such variability might explain the limited decrease in the error rate observed. As part of our future work, we plan to focus on developing methods that are robust in multi-writer (i.e., multi-target) scenarios.

\section{Ablation study with the BN layers}
\label{sec:bn_ablation}

\begin{figure}[!ht]
    \centering
    \includegraphics[width=1\textwidth]{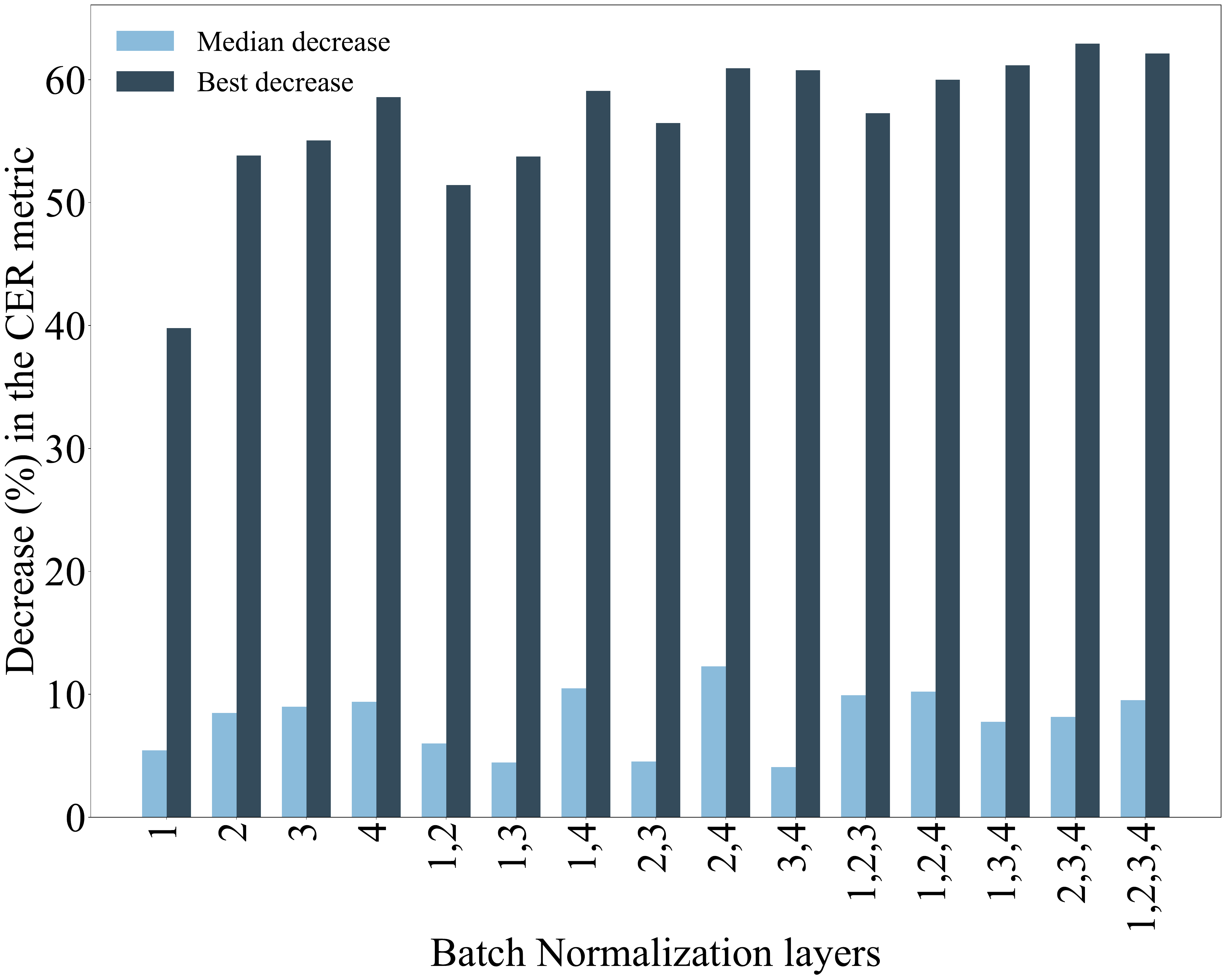}
    \caption{Impact of each possible combination of BN layers in the AMD adaptation. We report the median decrease in CER over the aggregation of the Top-1 result in each source-target configuration for a better notion of their impact. Additionally, we provide the best overall decrease achieved by each combination of BN layers. The CRNN architecture employed in this work includes four BN layers. A numerical identifier is used in order to distinguish the BN layer, starting from 0.}
    \label{fig:bn_ablation}
\end{figure}

As described in Sec. \textcolor{red}{3.3}, the Align loss introduced in AMD can be applied to either a single BN layer or a combination of them. In the latter case, an Align loss is calculated for each layer, with the final Align loss being the sum of all of them. In order to assess the particular impact of the selected combination of BN layers in the AMD adaptation, Fig.~\ref{fig:bn_ablation} presents the median and maximum decrease in the CER metric of this ablation study. The results indicate that fine-tuning based solely on the statistics of the first BN layer yields less favorable results. This might be because the initial convolutional layers extract features that are more general and suitable for a wide range of datasets, whereas as we progress deeper into the model, the features become increasingly specific to the characteristics of the training data. Thus, involving the statistics of the latter BN layers seems to yield the best results. In summary, and generally speaking, the deeper the BN layers considered for the Align loss, the better the outcomes.

\newpage

\end{appendix}

\bibliographystyle{unsrt}  
\bibliography{references}

\end{document}